# CFRecs: Counterfactual Recommendations on Real Estate User Listing Interaction Graphs

Seyedmasoud Mousavi
smousa13@asu.edu
Arizona State University
Tempe, Arizona, USA

Ruomeng Xu
ruomengx@zillowgroup.com
Zillow Group
Seattle, Washington, USA

Xiaojing Zhu
xiaojingz@zillowgroup.com
Zillow Group
Seattle, Washington, USA

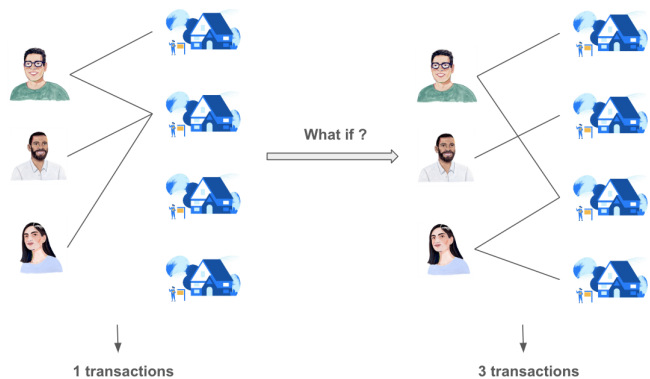

Figure 1: How would different circumstances that were not factually observed change the outcome?

## Abstract

Graph-structured data is ubiquitous and powerful in representing complex relationships in many online platforms. While graph neural networks (GNNs) are widely used to learn from such data, counterfactual graph learning has emerged as a promising approach to improve model interpretability. Counterfactual explanation research focuses on identifying a counterfactual graph that is similar to the original but leads to different predictions. These explanations optimize two objectives simultaneously: the sparsity of changes in the counterfactual graph and the validity of its predictions. Building on these qualitative optimization goals, this paper introduces CFRecs, a novel framework that transforms counterfactual explanations into actionable insights. CFRecs employs a two-stage architecture consisting of a graph neural network (GNN) and a graph variational auto-encoder (Graph-VAE) to strategically propose minimal yet high-impact changes in graph structure and node attributes to drive desirable outcomes in recommender systems. We apply CFRecs to Zillow's graph-structured data to deliver actionable recommendations for both home buyers and sellers with the goal of helping them navigate the competitive housing market and achieve their homeownership goals. Experimental results on Zillow's user-listing interaction data demonstrate the effectiveness of CFRecs, which also provides a fresh perspective on recommendations using counterfactual reasoning in graphs.

## CCS Concepts

• **Computing methodologies** → **Neural networks**.

## Keywords

graph-structured data, counterfactual graph learning, counterfactual explanation, recommender system, GNN, Graph-VAE

## 1 Introduction

In many real-world applications, data is collected from multiple entities that have complex relationships and interactions with each other. Graph structures are powerful tools for representing such heterogeneous complex networks [9]. Extracting valuable information from such graph data can benefit many domains and applications. One prominent example is recommendation systems [23], where Graph Neural Networks (GNNs) such as GCN [12], RGCN [17], GIN [24], and GAT [20] have been successfully used for learning patterns from complex network data.

Due to the lack of interpretability of such neural networks, they may only learn spurious correlations from data [13]. Counterfactual explanation has been used recently for promoting explainability and inferring causality on GNNs [1][3]. In graph counterfactual learning, the goal is to identify the necessary changes to alter the prediction of a learned Machine Learning model (such as a GNN) [21]. The factual evidence is the actual observed data that was also used to train the model. The generated counterfactual of a factual data point is therefore a similar data with a few changes in its structure [8][21]. In general, these explanations optimize two objectives simultaneously: the sparsity of changes in the counterfactual graph and the validity of its predictions, in order to identify a counterfactual graph that is similar to the original but leads to different predictions.

On the other hand, the increasing digitization of the real estate industry has led to an explosion of user-listing interaction data, enabling more personalized and efficient home discovery experiences. As a leading real estate marketplace, Zillow leverages these interactions to understand user preferences, home listing features, and their interdependencies. Such graph-structured data naturally captures the complex relationships, making GNN a powerful tool for modeling these interactions. While GNNs have demonstrated success in real estate recommendations, emerging research in counterfactual learning on graphs presents a novel opportunity to further enhance recommendation strategies by identifying actionable insights. Our research question is outlined in Figure 1: we would like to explore potential benefits of counterfactual learning in real estate data represented as graphs of interaction between users and listings.

Inspired by the works on GNN explanation, we propose CFRecs, a framework that optimizes the same objectives as GNN explanation models but generates counterfactual graphs that could be used for recommending actions. CFRecs is a two-stage model consisting of a graph neural network (GNN) and a graph variational autoencoder (Graph-VAE), which strategically generates minimal yet high-impact changes in graph structure and node attributes to drive desirable outcomes in recommender systems. This framework expands existing counterfactual graph generation to heterogeneous graphs and can be used for both regression and classification tasks.

In this paper, we make the following contributions:

- We outline a methodology for efficiently and robustly generating graph data points from a large body of interaction data, providing an essential set of graph data for model training and evaluation.
- We propose a novel framework for generating counterfactual recommendations in general user-item interaction graphs to achieve desirable outcomes.
- We conduct extensive experiments on Zillow's user-listing interaction data to demonstrate empirically the efficacy of generating recommendations using counterfactual reasoning in graphs.

## 2 Related Work

Several methods have been developed to generate counterfactual explanations for GNNs for various purposes such as assessing fairness and promoting interpretability [5]. They primarily focus on finding a subgraph that has a high correlation with the factual graph's prediction [25] and thus removing it will result in a changed GNN prediction.

RCExplainer [22] proposes explanations for the task of graph classification. By analyzing the decision regions of GNNs, RCExplainer leverages edge masks to generate robust counterfactual explanations while focusing on sparsity of changes and interpretability. However, it primarily considers edge removal and lacks feature modification capabilities. RCExplainer generates counterfactual edge masks that have the same prediction outcome as the factual graph while removing them from the original graph produces a different outcome.

CF-GNNExplainer [14] targets node-level counterfactual explanations by perturbing adjacency matrices. It optimizes binary mask matrices to change predictions for nodes, employing a sparsity regularization approach. Unlike RCExplainer, it does not seek to generate counterfactuals with desired labels but only ensures a prediction shift.

$CF^2$ [18] combines factual and counterfactual reasoning to produce explanations for graph classification that are necessary and sufficient. It uses dual constraints to ensure explanations retain sufficiency (reproducing predictions) and necessity (altering predictions upon removal), a feature distinguishing it from other frameworks that are focused only on one aspect.

CLEAR [16] employs a Variational Autoencoder (VAE) to generate counterfactual graphs, maintaining the latent causal structure of the input graph. By focusing on minimal changes to the graph structure and features, it ensures both interpretability and adherence to the original data distribution. This framework aligns closely with our goal of providing recommendations but still lacks the component of enforcing practical restrictions on graph changes for real estate recommendations.

There is no one-size-fits-all method for counterfactual learning. The approach heavily depends on the specific application, associated ML task, and optimization objective. To the best of our knowledge, it has not been applied in real estate recommendations. Therefore, we introduce a new framework tailored to this domain, addressing its unique challenges and objectives.

## 3 Problem Definition and Graph Data Creation

In this section, we provide a detailed description of the practical problem space our proposed framework aims to solve, along with a formal definition of counterfactual learning in this context and an illustration of how graph data is generated for learning.

### 3.1 Research Question

Given interactions (e.g., views, inquiries, and purchases) between users and home listings, how can we leverage counterfactual graph learning approaches to identify actionable insights that could increase the likelihood of successful home purchases and, consequently, lead to a higher total number of home transactions?

- Are we able to provide buyers with tailored suggestions to improve their chances of making a purchase, such as recommendations on engaging with properties they haven't considered and making adjustments to their budgets, location preferences, etc.?

- Are we able to strategically guide buyers to interact with different listings, thus discouraging bidding wars on homes and increasing the total number of successful home transactions?
- Are we able to provide sellers with suggestions on refining listings to help them sell properties faster?

One important factor to consider when forming those suggestions is the network effect. Changes made to a single node and edges may increase the likelihood of successful home purchases for one user, but could also result in a decrease for another user. To address this challenge, we aim to optimize the overall likelihood of transactions among the target user listing groups by considering the changes made on the graph as a whole.

As illustrated in the left part of diagram in Figure 1, the observed factual graph representing the actual user-listing interaction along with the features, this interaction network as a whole corresponds to a transaction count. Through counterfactual graph learning, we aim to know the transaction count from a counterfactual graph with an altered interaction network, i.e., "would the transaction count change if the graph structure or features became different from the observation?", as illustrated in the right part of the diagram. Additionally, we also aim to identify the counterfactual graph with minimal changes to the factual graph that could lead to the desired outcome.

There are also practical considerations to take into account when generating counterfactual graphs. The change is related to the feasible actions we can take from an application perspective. If the task is to figure out how to guide buyers to interact with different listings to increase total transactions, the counterfactual candidates we consider are only those perturbed graphs by adding and deleting edges from the original graph. In practice, there are several restrictions in adding or deleting edges when generating perturbed counterfactual graphs for what-if reasoning. For instance, if the listing is not within the price range for the user, then the corresponding edge cannot be added. As a result, graphs with such edges cannot be considered feasible counterfactual candidates unless there are alterations in the user's preferences, i.e., changes in the features along with changes in the graph structure for proposing a counterfactual candidate.

### 3.2 Counterfactual Learning

Formally, we define an interaction graph between users and listings as a tuple $\mathcal{G}$:

$$\mathcal{G} = (\mathcal{U}, \mathcal{L}, \mathcal{I}^{(1)}, ..., \mathcal{I}^{(m)}, \mathcal{X}^{(u)}, \mathcal{X}^{(l)}) \quad (1)$$

where $\mathcal{U}, \mathcal{L}$ are the sets of user nodes and listing nodes respectively with node characteristics $\mathcal{X}^{(u)}, \mathcal{X}^{(l)}$. $\mathcal{X}^{(l)}$ represents features describing a listing, such as the number of bedrooms, location, price, property type, year built, square footage, etc. $\mathcal{X}^{(u)}$ is a set of features capturing a user's preferences across different listing attributes based on their engagement history. The interactions between users and listings form a bipartite graph where the edges of multiple types of interaction are stored in adjacency matrices $\mathcal{I}^{(1)}, ..., \mathcal{I}^{(m)}$, where:

$$\mathcal{I}^{(i)}_{u,l} = \begin{cases} 1, & \text{if user } u \text{ has interaction of type } i \text{ with listing } l, \\ 0, & \text{otherwise.} \end{cases}$$

In our experiments, we worked with 3 adjacency matrices, one for each of the interactions types: views, saves, and submits.

A graph level label $y$ is defined as a binary variable to denote whether the graph will lead to a transaction in the future.

Given an interaction graph $\mathcal{G}$ between users and listings, the factual observations are the interactions, the user and listing node features, and the graph transaction label. A counterfactual of the graph $\mathcal{G}$ is defined as a tuple $\mathcal{G}'$:

$$\mathcal{G}' = (\mathcal{U}, \mathcal{L}, \mathcal{I}'^{(1)}, ..., \mathcal{I}'^{(m)}, \mathcal{X}'^{(u)}, \mathcal{X}'^{(l)}) \quad (2)$$

This counterfactual graph is defined on the same sets of users and listings but with different interactions, node features and graph label. Since the counterfactual graph $\mathcal{G}'$ is not observed, a label $y'$ is unavailable. However, using a neural graph classifier, we are able to estimate $y'$ with:

$$\hat{y}' = f(\mathcal{G}') \quad (3)$$

The counterfactual graph $\mathcal{G}'$ is generated in a way to satisfy multiple objectives [8]. The first is sparsity of changes. We desire $\mathcal{I}'^{(i)} \simeq \mathcal{I}^{(i)}$ to make the number of generated new interactions to be minimal. This is due to the fact that realizing the counterfactual graph in practice is not easy. We also desire $\mathcal{X}'^{(u)} \simeq \mathcal{X}^{(u)}, \mathcal{X}'^{(l)} \simeq \mathcal{X}^{(l)}$ for the same reason. For listing features, not all the attributes are mutable. For instance, number of bedrooms, number of bathrooms, square footage, etc. The only attribute that could change is the listing price. For users, user preference features are mutable, which can be interpreted as if the user were to adjust their home preferences.

The other objective is validity of the label. In other words, we want the counterfactual graph $\mathcal{G}'$ to have a higher chance of being transactional. We therefore desire $f(\mathcal{G}') > f(\mathcal{G})$. The two sets of objectives related to sparsity and validity require a trade off where the generated changes to the graph structure of $\mathcal{G}$ must be relevant. In other words, if a change to the graph structure does not lead to a higher chance of transactionality for $\mathcal{G}'$, then it is redundant and will be eliminated during the training process. This forces the counterfactual generation model to generate minimal changes that lead to a higher chance of transactionality.

### 3.3 Subgraph Creation

The graph of interactions formed from user engagements with listings spans across all the states and across multiple days. To learn the relationship between graph $\mathcal{G}$ and label $y$, we now propose an effective approach to obtain multiple subgraphs from a single large interaction graph along with their corresponding transaction labels.

There are many approaches to create subgraphs. One intuitive method is to use graph clustering algorithms [2]. However, these method usually result in one huge subgraph with many small isolated subgraphs. Another approach is to create subgraphs based on zip code. This approach has some drawbacks, including graph sizes and inability to generate interactions for listings outside a zip code, which is required for the application.

A simple alternative approach is random walking [15]. Starting from a random initial node (either a user or a listing), at each time step the walker moves to a random neighbor. The walker repeats this process until a certain number of nodes have been visited. If the walker reaches a leaf node, it goes back to one of the previous nodes. Finally, a subgraph is induced from the set of visited nodes. This approach creates subgraphs with more or less the same size and follows the natural flow of interactions between users and listings. The resulting subgraph might have listings from different zip codes which allows for generative interactions to other zip codes for users.

A simple random walk is a robust method to create subgraphs of interactions between users and listings. However, most of the subgraphs created with random walk are non-transactional. To make sure that a subgraph is transactional, only the first step needs to change. Inspired by [19], for a pair of transactional user listing nodes, random walk starts from either of the transactional nodes. This guarantees that a transactional subgraph is created. Algorithm 1 illustrates the process we followed for creating a subgraph from a graph of interactions, which provides an essential set of graph and transaction label data for the upcoming model training and evaluation.

**Algorithm 1** Random Walk in a Graph of Interactions Between Users and Listings

**Require:** $\mathcal{G} = (\mathcal{U}, \mathcal{L}, \mathcal{I}^{(1)}, ..., \mathcal{I}^{(m)}, \mathcal{X}^{(u)}, \mathcal{X}^{(l)})$
**Require:** $k \in \mathbb{N}$
**Ensure:** $P = [v_0, v_1, ..., v_k]$
1: $G \leftarrow$ undirected simple graph of $\mathcal{G}$
2: **if** $G$ has a transaction pair of nodes **then**
3:    $v_0 \leftarrow$ Either node of the transaction pair
4: **else**
5:    $v_0 \leftarrow$ Randomly select a node from $G$
6: **end if**
7: $P \leftarrow [v_0]$
8: $v \leftarrow v_0$
9: **for** $i \leftarrow 1$ to $k$ **do**
10:    $N(v) \leftarrow$ Neighbors of $v$ in $G$ not in $P$
11:    **if** $N(v) = \emptyset$ **then**
12:      $v \leftarrow$ Randomly select a vertex from $P$
13:    **else**
14:      $v \leftarrow$ Randomly select a vertex from $N(v)$
15:      $P \leftarrow P \cup [v]$
16:    **end if**
17: **end for**
18: **return** subgraph induced on $P$ from $\mathcal{G}$

## 4 Methodology

In this section, we describe the two key components of our CFRecs, which consists of a GNN for classification and a graph variational auto-encoder (Graph-VAE) for counterfactual graph generation.

### 4.1 Graph Classification

Since generated counterfactual graphs are not observed, their transactionality may only be estimated using a proxy. Given a dataset of observed factual graphs $\{\mathcal{G}^{(i)}\}$ with transaction labels $\{y^{(i)}\}$, a graph classifier is a function that learns to map each graph to its corresponding label. A neural graph classifier $f$ takes as input a graph of interactions and outputs a probability distribution to predict if the input graph is transactional.

We used a Graph Neural Network (GNN) as our neural graph classifier. Before graph message passing [6], node features need to be projected to the same space. A user $u$ has preferences for each listing attribute $a$, which are represented as a histogram of possible values $v_i$ and their corresponding frequencies $p_i$. The user has a total of $N_u^{(a)}$ preferences for each attribute $a$. These preferences can be expressed as:

$$x_u^{(a)} = \{v_1 : p_1, v_2 : p_2, ..., v_{N_u^{(a)}} : p_{N_u^{(a)}}\}$$

Attribute values are either categorical in nature such as zip code, or are binned to become categorical such as square footage. Users also have some real valued features denoting their activity within the website. Listing node features are both real valued and categorical.

The first layer of the graph classifier is node embedding. Given a total of $C$ categorical values for listing attributes, an embedding matrix $E \in \mathbb{R}^{C \times d_{emb}}$ is used to embed nodes into a $d_{emb}$-dimensional hidden space. For user nodes, a weighted average is used for each attribute. Given user preferences $x_u^{(a)}$, the hidden representation $h_u^{(a)}$ is obtained as:

$$h_u^{(a)} = \sum_{i=1}^{N_u^{(a)}} p_i E o_i \quad (4)$$

where $o_i$ is the one-hot vector where the entry associated with $v_i$ is one, and all other entries are zero. For each user, preferences over all the attributes are computed and concatenated together with real valued features to obtain user embeddings $h_u$. Listing embeddings $h_l$ are obtained simply without the need for averaging.

Once user and listing node embeddings are obtained, two linear transformations are used to encode node embeddings $h$ into a $d$-dimensional space denoted as $z \in \mathbb{R}^d$. The input to the GNN is a tuple $(Z, \mathcal{I}^{(1)}, \mathcal{I}^{(2)}, ..., \mathcal{I}^{(m)})$, where $Z \in \mathbb{R}^{n \times d}$ has $n$ rows for each of the $n$ nodes. We used the methodology proposed in [12] for message passing of each interaction type. Interaction adjacency matrices $\mathcal{I}^{(i)}$ are normalized to obtain $\mathcal{J}^{(i)}$ as:

$$\mathcal{J}^{(i)} = \tilde{D}_i^{-\frac{1}{2}} \tilde{\mathcal{I}}^{(i)} \tilde{D}_i^{-\frac{1}{2}} \quad (5)$$

where $\tilde{\mathcal{I}}^{(i)} = \mathcal{I}^{(i)} + I_n$, with $I_n$ being the identity matrix, and $\tilde{D}_i$ is the diagonal degree matrix obtained from $\tilde{\mathcal{I}}^{(i)}$. To allow for a better message passing, another interaction adjacency matrix is added to densify the graph:

$$\mathcal{I}^{(m+1)} = \mathcal{I}^{(i)} + \mathcal{I}^{(i)^2} \quad (6)$$

where the $m+1$'th interaction adjacency matrix is obtained from the $i$'th interaction. We chose the first interaction type (views) as it was the densest interaction type.

We used a relational GNN similar to [17] to aggregate information across all interaction types. Graph message passing is performed by the GNN to obtain salient node embeddings by aggregating local neighborhood information. Each layer of GNN aggregates node features $Z^{(l)}$ from one hop of neighborhood away with learnable weight matrices $W_j^{(l)}$ for each layer $l$ and each interaction type $j$ and non linearity function $\sigma$:

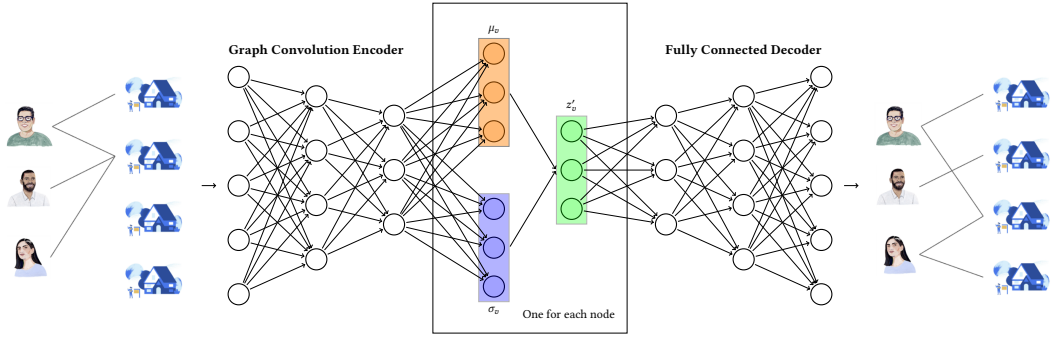

Figure 2: Counterfactual graph generator architecture. The Graph Convolution Encoder learns node distribution vectors for each node. The Fully Connected Decoder transforms these vectors to refine the node level and edge level graph structure.

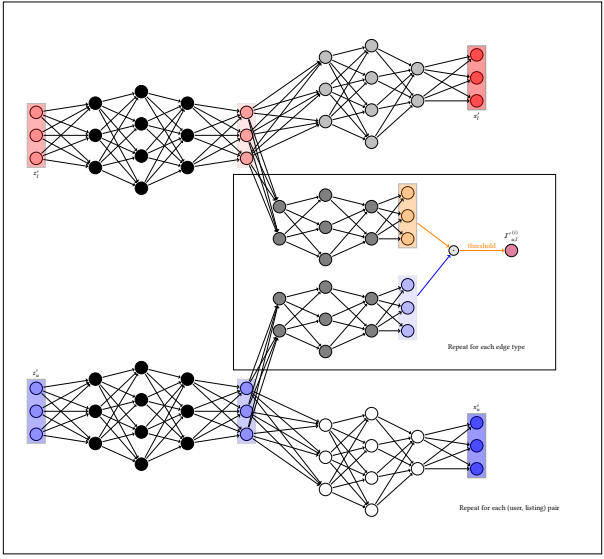

Figure 3: Architecture of the fully connected decoder. This component is used to generate the links and node features of the counterfactual graph. Inner nodes are colored to indicate weight sharing.

$$Z^{(l+1)} = \sigma(\sum_{j=1}^{m+1} \mathcal{J}^{(j)} Z^{(l)} W_j^{(l)}) \quad (7)$$

After L layers of GNN message passing, $Z^{(L)}$ provides node representations. For graph classification, element-wise global graph pooling is applied to obtain one feature vector $z^{(graph)}$. This vector is then fed into a Multi Layer Perceptron (MLP) with the last layer outputing $\hat{y} \in [0, 1]$. Weight matrices and the embedding matrix $E$ are learned by gradient descent to minimize the Binary Cross Entropy (BCE) loss over N graphs:

$$\mathcal{L} = -\frac{1}{N} \sum_{i=1}^{N} [y_i \log(\hat{y}_i) + (1 - y_i) \log(1 - \hat{y}_i)] \quad (8)$$

### 4.2 Counterfactual Graph Generation

After the GNN is trained to classify graphs as transactional or non-transactional, it is used as an oracle to estimate the transactionality of any counterfactual unobserved graph. We used a variational graph auto-encoder architecture [11] for counterfactual graph generation. Figure 2 depicts the architecture of the counterfactual graph generation model. The encoder is comprised of a GNN that learns node distributions. The weights of the GNN encoder could be initialized with the weights of the GNN classifier before the last classification layers to make use of transfer learning [26]. The same architecture including the node embedding techniques used for the graph classifier are used for the encoder. After L layers of GNN message passing, $Z^{(L)}$ is fed into two MLPs, one to obtain mean vectors $\mu_i$ and one to obtain diagonal covariance matrices, represented as vectors $\sigma_i$ for each node $i$.

The decoder samples from node distributions $\mathcal{N}(\mu_i, \sigma_i^2)$ with the reparameterization trick [10] to obtain a random dense feature vector $z'_i$. Two MLPs are used to learn user and listing features $x'^{(u)}$ and $x'^{(l)}_{price}$. The MLP for listings only learns price, and the MLP for users learns user preference histograms. 3 more MLPs are also used to transform $z'_i$ vectors for each interaction type. The edges of the generated graph for each interaction type are obtained by a similarity score as :

$$\mathcal{I}'^{(i)}_{j,k} = \text{sigmoid}(<g^{(i)}(z'_j), g^{(i)}(z'_k)>) \quad (9)$$

where $g^{(i)}$ is the MLP for interaction i and $<,>$ is the dot product operator. Figure 3 illustrates the architecture of the decoder, which consists of multiple fully connected neural components.

The parameters of the counterfactual generator are learned by optimizing a loss function consisting of multiple objective functions. For node features, the objective function is defined for $n_1$ listings and $n_2$ users as:

$$\mathcal{L}_X = \sum_{l=1}^{n_1} ||x^{(l)}_{price} - x'^{(l)}_{price}||_2^2 + \sum_{u=1}^{n_2} ||x^{(u)} - x'^{(u)}||_2^2 \quad (10)$$

For the interactions, the objective function is defined as:

$$\mathcal{L}_\mathcal{I} = \sum_{i=1}^{m} \sum_{j=1}^{n} \sum_{k=1}^{n} \alpha_i BCE(\mathcal{I}^{(i)}_{j,k}, \mathcal{I}'^{(i)}_{j,k}) \quad (11)$$

Table 1: Comparison of Counterfactual Generator Performance Under Different Training Settings

| Views Added (%) | Saves Added (%) | Submits Added (%) | Views Removed (%) | Saves Removed (%) | Submits Removed (%) | User Preferences Similarity (%) | Listing Price Similarity (%) | Average Lift (%) | Total Increase (%) | $(\gamma, \zeta, \beta, \eta)$ | Figure |
|---|---|---|---|---|---|---|---|---|---|---|---|
| 18.63 | 17.29 | 40.70 | 0.78 | 2.31 | 5.67 | 50.78 | 99.15 | 7.2 | 83.0 | (0.2, 1, 1, 1) | 4 |
| 18.63 | 0.00 | 0.00 | 0.78 | 0.00 | 0.00 | 100.00 | 100.00 | 2.6 | 76.3 | (0.2, 1, 1, 1) | 5b |
| 34.98 | 0.00 | 0.00 | 0.78 | 0.00 | 0.00 | 100.00 | 100.00 | 3.7 | 78.5 | (0.2, 1, 1, 1) | 5a |
| 6.04 | 0.00 | 0.00 | 0.79 | 0.00 | 0.00 | 100.00 | 100.00 | 1.0 | 60.0 | (0.2, 1, 1, 1) | 5c |
| 5.47 | 0.00 | 0.00 | 0.81 | 0.00 | 0.00 | 100.00 | 100.00 | 2.4 | 69.8 | (0.4, 0.5, 0.5, 2) | 6b |
| 21.58 | 0.00 | 0.00 | 0.80 | 0.00 | 0.00 | 100.00 | 100.00 | 3.8 | 76.4 | (0.4, 0.5, 0.5, 2) | 6a |

The coefficients $\alpha_i$ are chosen to emphasize which interaction type is harder to realize in practice. For instance, by choosing $\alpha_3$ to be higher, we force the model to generate fewer new interactions of type 3 because realizing it in practice is less feasible.

The newly generated graph once fed into the graph classifier gives an estimate of its transactionality. Our objective is to increase the chances of transactionality by a margin $\gamma$. To reflect this, we set this loss for transactional graphs to be zero and define the counterfactual objective function over non-transactional graphs $\mathcal{G}$ as a margin-based ranking loss:

$$\mathcal{L}_{CF} = \max(0, f(\mathcal{G}) - f(\mathcal{G}') + \gamma) \quad (12)$$

To ensure that the node distributions are Gaussian, we also add a KL divergence loss for n nodes as:

$$\mathcal{L}_{KL} = \frac{1}{n}\sum_{i=1}^{n} KL(\mathcal{N}(\mu_i, \sigma_i^2), \mathcal{N}(0, I)) \quad (13)$$

The full loss function is then:

$$\mathcal{L} = \zeta \mathcal{L}_I + \beta \mathcal{L}_X + \eta \mathcal{L}_{CF} + \lambda \mathcal{L}_{KL} \quad (14)$$

The coefficients $\zeta, \beta, \eta, \lambda$ are chosen to guide the model to balance which objectives are more important.

## 5 Experiments and Results
## 5.1 Graph Classification Performance

To train the oracle of graph classification we used a dataset of 7946 graphs evenly divided between transactional and non-transactional Zillow interaction data, which were obtained through a random walk described in section 3.3. We used 6946 data points for training and 500 data points for each of the test and validation sets. We regularized the GNN classifier with dropout layers and using early stopping achieved the results in Table 2. The ROC-AUC for the test data is 0.71, demonstrating an reasonable classification performance.

Table 2: Graph classification ROC AUC scores for different dataset splits.

| Dataset | ROC AUC (%) |
|---|---|
| Validation | 74.789 |
| Test | 71.284 |
| Training | 72.961 |

## 5.2 Counterfactual Graph Generation Performance

For the main task of counterfactual graph generation we used a smaller subset of 2583 graphs evenly divided between transactional and non-transactional graphs. 250 graphs are for testing, 250 graphs are for validation, and the rest are for training. A random baseline is used to compare the performance of the counterfactual generator. The random model generates the same number of edge changes as the counterfactual model.

To evaluate the performance of the generated counterfactual graph, we compare two types of evaluation metrics: proximity-based and validity-based.

- Proximity measures the similarity between the generated counterfactual graph and the original graph. To assess this, we report the percentage of edges added and deleted for each of the three types of interaction graphs compared to the original graphs, as well as the cosine similarity between the counterfactual and original node features for both user and listing features.
- Validity measures how effectively the counterfactual graph alters the outcome to the desired value. To evaluate this, we report the average percentage lift in transaction probability compared to the original interaction graph, as well as the percentage of counterfactual graphs exhibiting a positive lift relative to the original. For a direct comparison against the random baseline, we also present the relative lift of the Graph-VAE generated counterfactul graph compared to a randomly generated one.

*5.2.1 Unconstrained Model Performance.* The first experiment concerns the performance of the model in its full setting without any restriction on proposed graph changes. In this scenario, node features, user preferences and listing prices, and all edge types are generated by the counterfactual graph generation model. Edges may be added to work as recommendations or removed to allow for comparison of different recommendation models as if some edges are redundant. The model generates a number between 0 and 1 for each pair of user and listing and for each edge type. If a link has a value higher than some threshold, it is added to the graph, and if it is below some other threshold it is removed if it exists in the original graph. By tuning these two thresholds, one is able to control the degree of changes to the input graph. Table 1 summarizes the results of different experiments.

Figure 4 illustrates the performance of the model in its full setting. Overall, 83% of generated graphs have higher transaction probability, with an average of 7.2% lift. This is achieved by an average of 18.63% more views, 17.29% more saves, and 40.7% more submits. The latter is skewed as most graphs have no submit links and the model generating a few submits pushes the average up. The model learns to generate listing prices very closely to the inputs, but less so for user preferences as the data it is trained on is gathered from a shorter period of time than the original source data. As shown in Figure 4, the counterfactual graphs generated have significantly higher transaction probability than the random baseline generator. The random baseline's lift in transaction probability distribution is centered around zero signaling pure stochasticity.

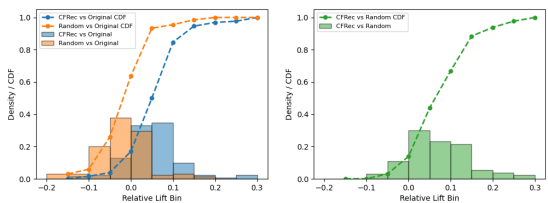

Figure 4: Left: Distribution of relative lift in transaction probability from CFRecs-generated graphs compared to the original interaction graphs. Right: Distribution of relative lift in transaction probability from CFRecs-generated graphs compared to randomly generated graphs.

*5.2.2 Constrained Model Performance.* The second set of experiments concerns the constrained model where only views are generated. Node features and saves/submits are fixed as in the input graph. Figures 5a, 5b, and 5c illustrate how different degrees of freedom for the generator impact the performance of the model. For this experiment, different thresholds were used to achieve different sparsity levels of change. Intuitively, more interactions in the graph should lead to a higher transaction likelihood. This is empirically observed as an increase of roughly 6% in views leads to an average of only 1% lift in transaction probability with around 60% of graphs seeing lifts, whereas an increase of roughly 18% in views leads to 2.6% average lift with around 78% of graphs experiencing lifts.

*5.2.3 Performance Under Different Tuning Parameters.* The third experiment concerns the effects of changing the coefficients for different loss terms during training of the counterfactual generator. The model discussed in the first two experiments used the same coefficients. A margin of 0.2 was used for the counterfactual loss function. The coefficients for the edge and feature reconstructions and the counterfactual loss were set to 1. Another model was also trained with a heavy focus on the counterfactual loss term. The margin was increased to 0.4 so the model is forced to generate graphs with higher level of transaction likelihood. The loss coefficients for edge and feature reconstructions were reduced to 0.5 while at the same time the coefficient for the counterfactual term was doubled to 2. As shown in Figures 6b and 6a, with a similar level of sparsity around 6%, the model trained with a higher level of transaction likelihood in mind generates graphs with higher lifts on average (2.4% vs. 1% on average) where also more graphs are generated with lifts (69.8% vs. 60%). Another interesting observation as illustrated in Figures 6b and 6a is that the second model generates graphs with slightly higher average lifts (3.8% vs. 3.7%) with almost 40% fewer edges added (21.58% vs. 34.98%) where almost generating the same number of graphs with lifts.

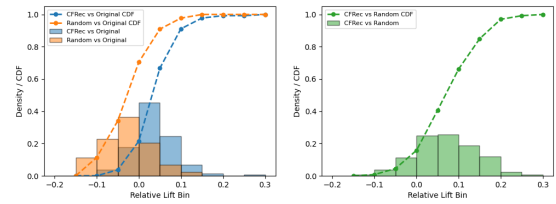

(a) Only +34% more views

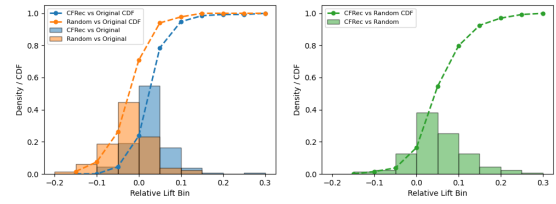

(b) Only +18% more views

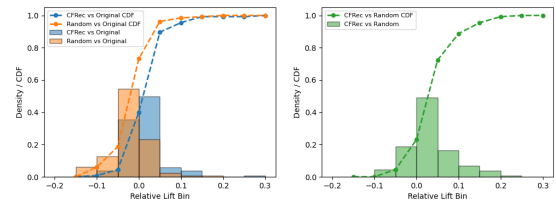

(c) Only +6% more views

Figure 5: Distribution of relative lift in transaction probability from CFRecs generated graphs compared to the original interaction graphs (left), and compared to the randomly generated graphs (right), under the constrained settings: only views can be changed. Different thresholds were picked to add an edge to the graph. In (a), a lower threshold allows for addition of more edges which in turn yields a higher likelihood of transaction. In (c), the model is restricted to add only a small number of edges and the distribution shows slim lifts in transaction probability. A balance between (a) and (c) is depicted in (b) to illustrate our intuition that more engagement naturally improves transactionality.

Figure 7 illustrates how a generated counterfactual graph could be used in a recommendation system. The counterfactual model generates 12% more view edges, and the GNN oracle predicts a 6.5% increase in the chances of transaction for the graph. Factual edges are shown in gray, and the counterfactual edges are shown in green. Noteworthy is the fact that due to the stochasticity associated with the generative process, one might obtain several graphs generated from the same input graph and choose to implement the graph with the highest lift in transactionality or the lowest implementation cost.

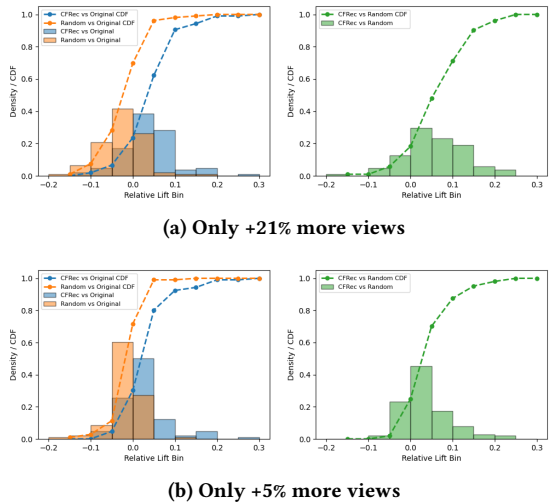

(a) Only +21% more views

(b) Only +5% more views

Figure 6: Distribution of relative lift in transaction probability from CFRecs generated graphs compared to the original interaction graphs (left), and compared to the randomly generated graphs (right), under the constrained settings: only views can be changed, and a different tuning parameter setting: counterfactual margin 0.4, training coefficients (0.5, 0.5, 2).

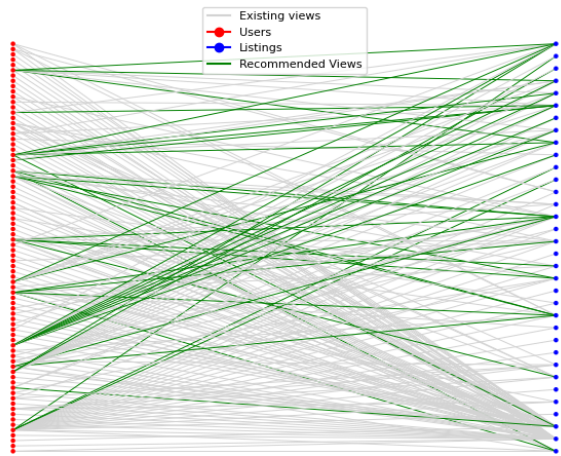

Figure 7: Illustration of how a counterfactual graph with only 12% more views could potentially have a 6.5% lift in transaction probability in a user listing interaction graph.

## 6 Conclusion and Future Work

In this work we demonstrated how counterfactual graph learning could be used to generate recommendations in a real-world enterprise application. This work provides a framework that could easily be adopted for any user-item interaction data with different structures, interactions, and desired output signal. This framework allows the user to constrain the generation of recommendations to take into account the costs of implementing each type of recommendation in practice.

We demonstrated that the objectives of counterfactual learning not only are applicable to explaining the predictions of classification models, but also generalizable to such applications as recommendation systems.

We showed through multiple experiments that the generated counterfactual interaction graphs could potentially yield better outcomes if implemented. Due to the unavailability of ground truth data to evaluate the counterfactual graphs, one could only trust the evaluation obtained from the GNN classifier. However, the GNN classifier is not 100% accurate, and is only an estimation of actual outcomes.

There are multiple directions for further exploration in the future. The first and foremost being deployment in production. Using counterfactual explanations to inform real estate recommendations is a novel idea and involves the development of a brand-new product. This product not only provides home recommendations to buyers but also suggests actionable steps for buyers, such as connecting with an agent, adjusting their offer, or exploring alternative locations and price ranges. For sellers, it offers recommendations like reducing the listing price, renovating the kitchen, or delaying the listing by a month. The integration of all these components will require the creation of a new product. The deployment to live users will require collaboration across multiple teams and will take time to materialize. Additionally, online A/B testing must account for network effects, as restricting users from viewing certain homes would be unethical and potentially violate fair housing law. Therefore, a specially designed experiment is necessary.

Another direction for future work is utilizing more sophisticated search-based methods for counterfactual generation. Reinforcement Learning [7] or GFlowNets [4] are promising directions to follow in the future.

To make the generated counterfactual more trustworthy, another direction would be to conduct more rigorous training and evaluation of the GNN classification oracle, and to derive the lower bounds of GNN accuracy for a certain minimum level of trustworthiness. More advanced GNN models exist whose application to the classification of transactional graphs should potentially yield better accuracies.